\documentclass[10pt,twocolumn,letterpaper]{article}

\usepackage{cvpr}
\usepackage{times}
\usepackage{epsfig}
\usepackage{graphicx, multirow}
\usepackage{amsmath}
\usepackage{amssymb}

\usepackage[breaklinks=true,bookmarks=false]{hyperref}

\cvprfinalcopy % *** Uncomment this line for the final submission

\def\cvprPaperID{2457} % *** Enter the CVPR Paper ID here

\ifcvprfinal\fi
\begin{document}

%%%%%%%%% TITLE
\title{Efficient and Deep Person Re-Identification using Multi-Level Similarity}

\author{Yiluan Guo, Ngai-Man Cheung\\
	ST Electronics - SUTD Cyber Security Laboratory\\
	Singapore University of Technology and Design\\
	\tt\small yiluan$\_$guo@mymail.sutd.edu.sg, \tt\small ngaiman$\_$cheung@sutd.edu.sg	
	% For a paper whose authors are all at the same institution,
	% omit the following lines up until the closing ``}''.
	% Additional authors and addresses can be added with ``\and'',
	% just like the second author.
	% To save space, use either the email address or home page, not both
	%\and
	%Ngai Man Cheung\\
	%Institution2\\
	%First line of institution2 address\\
	%{\tt\small ngaiman$\_$cheung@sutd.edu.sg}\\	 
	%{Singapore University of Technology and Design}
}

\maketitle
%\thispagestyle{empty}

%%%%%%%%% ABSTRACT
\begin{abstract}
Person Re-Identification (ReID) requires comparing  two  images of person  captured under different conditions. Existing work based on neural networks often computes the similarity of feature maps from one single convolutional layer. In this work, we propose an efficient, end-to-end fully convolutional Siamese network that computes the similarities at multiple levels. We demonstrate that multi-level similarity can improve the accuracy considerably using low-complexity network structures in ReID problem. Specifically, first, we use several convolutional layers to extract the features of two input images. Then, we propose Convolution Similarity Network to compute the similarity score maps for the inputs. We use spatial transformer networks (STNs) to determine spatial attention. We propose to apply efficient depth-wise convolution to compute the similarity. The proposed Convolution Similarity Networks can be inserted into different convolutional layers to extract visual similarities at different levels. Furthermore, we use an improved ranking loss to further improve the performance. Our work is the first to propose to compute visual similarities at low, middle and high levels for ReID. With extensive experiments and analysis, we demonstrate that our system, compact yet effective, can achieve competitive results with much smaller model size and computational complexity.
\end{abstract}

%%%%%%%%% BODY TEXT
\section{Introduction}

In person re-identification (ReID), given one image for a particular person captured by one camera, we need to re-identify this person from multiple images in a gallery captured by different cameras from different viewpoints. This task has attracted much attention due to its various applications in video surveillance and image retrieval. Substantial works have been proposed to accomplish this task, but improvement is  still needed. The main challenge is the significant visual appearance changes caused by the illumination variation, occlusion, viewpoint change, person pose as well as background clutter. 
Also, the computation needs to be efficient to handle large volumes of surveillance videos.

In existing works, Person ReID is solved from two perspectives. One is to develop a powerful representation to discriminate different identities \cite{xiong2014person,zhao2013person,li2013locally} and the other is to design an effective distance metric so that the similarity between different images can be measured \cite{Chen_2015_CVPR,chen2014relevance,guillaumin2009you}. 

With the great success achieved by deep convolutional nets (ConvNets) in computer vision \cite{krizhevsky2012imagenet,Szegedy_2015_CVPR,He_2016_CVPR,Toan:2016,Hoang:2017}, some works have been proposed to address Person ReID with deep neural networks in an end-to-end fashion. One approach is to classify the images into different identities \cite{Chen_2017_ICCV_Workshops,Li_2017_CVPR,Su_2017_ICCV}. During testing, each image is represented by the output of final fully connected layer before the classifier. This approach may suffer from the fact that, in ReID, there are only limited images for each identity during training. \cite{li2014deepreid} formulated Person ReID as a binary classification problem with Siamese network structure. Two images are fed into the network that determines whether they are matched or not. This approach alleviated the problem of insufficient training samples and achieved state-of-the-art results at that time. The critical component of this formulation is how to measure the similarity of two input images.  \cite{li2014deepreid} measured the similarity by computing the product of horizontal stripes. \cite{Ahmed_2015_CVPR} extended this idea by taking the neighborhood into consideration. The similarity is computed as the difference between one pixel on feature maps for one image and its $5\times5$ neighbors on feature maps for another image. \cite{subramaniam2016deep} further enlarged the neighbor search area and computed the correlation as the similarity.

\begin{figure}[htb]
	\centering
	\begin{minipage}[]{.3\linewidth}
		%\centering
		\centerline{\includegraphics[width=2.0cm]{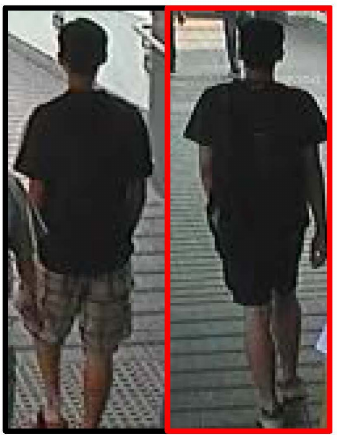}}
		\vspace{0.1cm}
		\centerline{(a)}\medskip
		\label{fig1a}
	\end{minipage}
	%\hfill
	\hspace{0.01cm}
	\begin{minipage}[]{.3\linewidth}
		%\centering
		\centerline{\includegraphics[width=2.0cm]{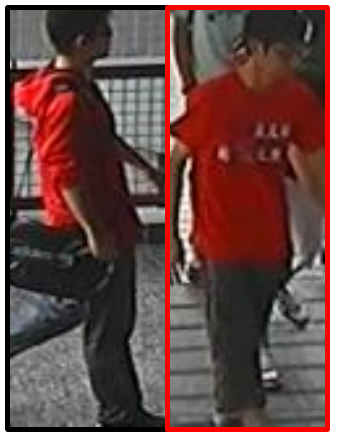}}
		%  \fbox{\includegraphics[width=4.0cm]{2}}
		\vspace{0.1cm}
		\centerline{(b)}\medskip
		\label{fig1b}
	\end{minipage}
	%\hfill
	\hspace{0.01cm}
	\begin{minipage}[]{.3\linewidth}
		%\centering
		\centerline{\includegraphics[width=2.0cm]{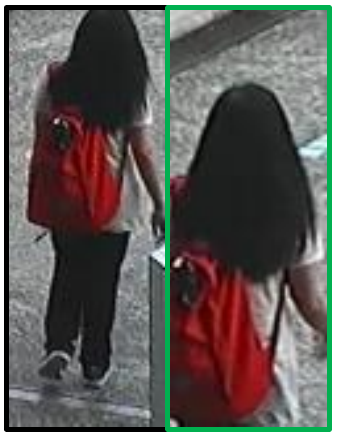}}
		%  \fbox{\includegraphics[width=4.0cm]{2}}
		\vspace{0.1cm}
		\centerline{(c)}\medskip
		\label{figc}
	\end{minipage}
	\caption{Some examples that are difficult for existing methods. The image in red box indicates it is unmatched and green for matched. (a) and (b) are two unmatched pairs with quite similar appearance. The images in (a) and (b) are different in low level and high level visual features respectively. The red bag in (c) translates for a large distance, which makes the similarity computation difficult.}
	\label{fig1}
\end{figure}

All these methods have some limitations. First, they only consider the similarity for the outputs of a certain convolutional layer. However, we argue that for ReID, it is useful to use {\em multi-level similarity}, i.e., similarity of the features of the bottom convolutional layers that contain low level visual information, and that of the higher layers that contain semantical information. Figure \ref{fig1} contains some illustrative examples. Figure \ref{fig1}(a) and (b) are both non-matching pairs. For Figure \ref{fig1}(a), similarity based on low-level features that capture the color of the shorts can identify the difference of the two persons. For Figure \ref{fig1}(b), the cloth of the two persons is quite similar and low-level features would fail to distinguish the persons in this case. On the other hand, similarity computed from high-level features can indicate that one is carrying a bag and the other is not. Figure \ref{fig1}(c) is a matching pair. In this case, low and high-level features can be used simultaneously to identify that red color backpacks are present in both images, indicating that there is a high probability that the same person is captured in both images. Second, some previous works assumed that the visual features would not translate for a large distance, thus the computing of product and difference only considering the counterparts on the same or neighboring locations. Although \cite{subramaniam2016deep} enlarged the search region for computing correlation, it still failed to incorporate the possible correlation for the entire scale, which might lead to missing information. For example, the red color backpack in the two images of Figure \ref{fig1}(c) translates for a substantial distance. Third, in previous works, both the product and difference are computed between the rigid parts from feature maps, which are not invariant to scale, rotation. They are inadequate to handle the case when the two matched images are captured by two cameras with similar angle but from different distances.

In this paper, we propose  a fully-convolutional, Siamese network based design to address these issues. First, we compute the visual similarities at different levels of the whole network. Second, given that convolution can be viewed as the computation of ``correlation" between a filter and the signal, we formulate the computation of similarity between two images as the convolution between the extracted part from one image (filter) and the other whole image (signal). In this case, the computation is not restricted to a specific search window, addressing the issue of large translation distance. Third, by leveraging Spatial Transformer Networks (STN) \cite{jaderberg2015spatial}, we extract meaningful parts from the feature maps of the image. In this case, STNs introduce an attention mechanism into our system. Our contributions are: 
\begin{itemize}
	\item We propose a fully convolutional Siamese network for Person ReID. A new module named Convolution Similarity Network is proposed to improve the measurement of similarity between two input images. This new module exploits the attention mechanism and could be implemented efficiently.   
	\item We compute visual similarities at different levels and combine them to achieve robust matching/non-matching classification. 
	\item We conduct extensive experiments and show that our method achieves competitive results in comparison with state-of-the-arts, with a lower computational complexity and model memory.
\end{itemize}

%-------------------------------------------------------------------------
\section{Related Work}
Most existing methods for Person ReID could be divided into two classes, one is traditional methods and the other is deep learning based approaches. For the traditional methods, there are usually two stages. First, handcrafted features are computed such as color histogram \cite{xiong2014person,koestinger2012large}, Gabor features \cite{li2013locally} and dense SIFT \cite{zhao2013person}.The handcrafted features are expected to contain as much as possible discriminative information for different persons. Following stage is similarity metric learning. Different metric learning approaches have been proposed to decide whether two images are matched or not \cite{Chen_2015_CVPR,chen2014relevance,guillaumin2009you,li2013learning}. A suitable metric should indicate the similarity of two images based on the handcrafted features, \ie, two images for the same person should have a smaller distance than those for different persons. Some work even adopted an ensemble of different metrics \cite{paisitkriangkrai2015learning}. Since the feature extraction stage and metric learning are two independent components in traditional methods, the optimization of features and distance metric might not help each other and eventually become sub-optimal. Our proposed method is significantly different from all these traditional ones as we jointly learn the features and metric in a deep neural network.

On the other hand, deep ConvNets have achieved great success in computer vision tasks like object recognition, detection, semantic segmentation \cite{krizhevsky2012imagenet,ren2015faster,long2015fully}. Recently, some published work show the promising power of deep ConvNets in person ReID. \cite{li2014deepreid} proposed a Siamese network that takes a pair of images to be compared. Convolutional layers are used to extracted visual features and product is used to indicate the similarity. \cite{Ahmed_2015_CVPR} proposed an improved architecture where neighbor difference were used to measure the similarity. \cite{subramaniam2016deep} further extends this architecture by enlarging the neighbor search region and normalize the elements before computing product. All the above works formulate the Person ReID task as a binary classification problem. The difference between our work is that we leverage the spatial attention and integrate the computation of similarities at different levels into the fully convolutional structure. Meanwhile, there is another line of approaches formulating this task as a ranking problem \cite{Chen_2017_CVPR,Cheng_2016_CVPR,varior2016gated,Zhao_2017_ICCV}. There are two or three images as input. Contrastive loss or triplet loss are used to push the images for the same identity closer together and pull the images for different identities more far away in the embedding space. More recently, the combination of classification and ranking loss obtained promising results by taking advantage of both ranking and binary classification tasks \cite{Wang_2016_CVPR,chen2017multi}. Our method also adopts multiple tasks to train the network with the difference that the ranking loss is based on the attended regions extracted by STNs instead of the descriptors of the whole images because the meaningful parts of images excluding noise and redundancy are more effective to represent the identities.

Another interesting approach treats ReID as recognition problem and classifies the images to different identities directly. \cite{Zhao_2017_CVPR,Su_2017_ICCV} extracted different body parts by human pose and combined local and global features for classification. \cite{Chen_2017_ICCV_Workshops,Qian_2017_ICCV}, on the other hand, considered the features at different scale. Among these works, \cite{Li_2017_CVPR} proposing to use STNs to find the meaningful local parts is similar to ours. However, our method has a different goal for the usage of STNs: we want to compute the similarity explicitly. The network structure is also distinct since we build a Siamese network and the final object is binary classification.

\cite{Sun_2015_CVPR} exploited the idea of visual similarities on multi-level for face recognition, whereas our approach is proposed for the task of ReID. Moreover, we introduce attention mechanism and improve the similarity computation, which make the multi-level similarity more accurate and effective.
\section{Proposed Method}

\subsection{Model description}

\begin{figure}[htb]
	\centering
	\centerline{\includegraphics[width=8.5cm]{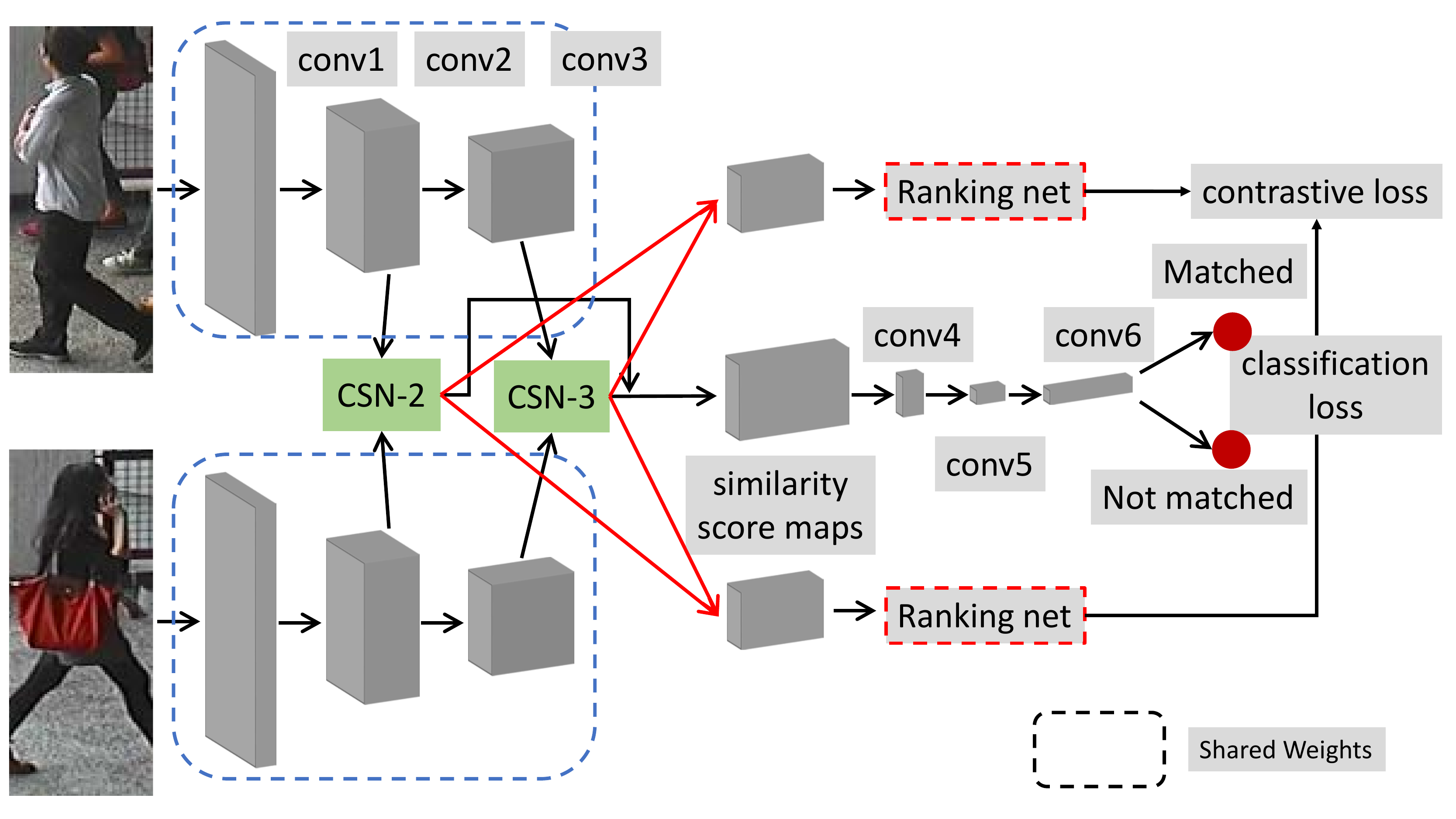}}
	\caption{The structure of proposed method. The outputs of CSN-2 and CSN-3 indicated by black arrows are concatenated together for further processing. The outputs of CSN-2 and CSN-3 indicated by red arrows are fed into the ranking net. Networks in the dash line boxes with the same color means that they are sharing the same parameters. Details of the CSN can be found in Figure \ref{fig3} and related context.
	}
	\label{fig2}
\end{figure}

The overall structure of the proposed model is shown in Figure \ref{fig2}.
Two input images $I_1$ and $I_2$ ($60\times160\times3$ in our experiments) are processed by three successive convolution layers. Let $x_i^{(j)}$ denote the output of the $j$-th convolution layer for $I_i$. The output of the second and third convolution layers, $x_1^{(2)}$,  $x_2^{(2)}$ as well as $x_1^{(3)}$, $x_2^{(3)}$ are fed into two Convolution Similarity Networks (CSNs), which have two sets of outputs: one is the similarity score maps for $I_1$ and $I_2$ while the other is feature maps for the extracted local parts of $I_1$ and $I_2$. The similarity score maps are processed by three more convolution layers. The details of convolutional layer 1-6 are listed in Table \ref{tab1}. Two nodes in the final layer indicate whether $I_1$ and $I_2$ are matched or not, i.e., binary classification. An additional objective function is to make the matched pair closer and unmatched pair far away in an embedding space. 
%In this section, we will introduce proposed CSN in detail first. Then we will describe the combination of two objective functions. 

\begin{table}[]
	\centering
	\footnotesize
	\caption{Network specifications. All the convolutional layers except the conv4, conv6 and loc conv3 are followed by $2\times2$ maxpooling. The conv6 and loc conv3
	are followed by 
	global average pooling (GAP) layers\cite{lin2013network}.}
	\label{tab1}
	\begin{tabular}{|c|c|c|c|}
		\hline
		Network                                                                     & Layer & filter size & \#filters \\ \hline
		\multirow{6}{*}{\begin{tabular}[c]{@{}c@{}}Whole\\ structure\end{tabular}}  & conv1 & $5\times5\times3$       & 32        \\ \cline{2-4} 
		& conv2 & $3\times3\times32$      & 96        \\ \cline{2-4} 
		& conv3 & $3\times3\times96$      & 96        \\ \cline{2-4} 
		& conv4 & $1\times1\times1152$    & 32        \\ \cline{2-4} 
		& conv5 & $3\times3\times32$      & 32        \\ \cline{2-4} 
		& conv6 & $1\times1\times32$      & 500       \\ \hline
		\multirow{3}{*}{\begin{tabular}[c]{@{}c@{}}Localization\\ Net\end{tabular}} & loc conv1 & $3\times3\times96$      & 32        \\ \cline{2-4} 
		& loc conv2 & $3\times3\times32$      & 32        \\ \cline{2-4} 
		& loc conv3 & $1\times1\times32$      & 128       \\ \hline
	\end{tabular}
\end{table}

%\subsection{Convolution similarity network}

\begin{figure}[htb]
	\centering
	\centerline{\includegraphics[width=8cm]{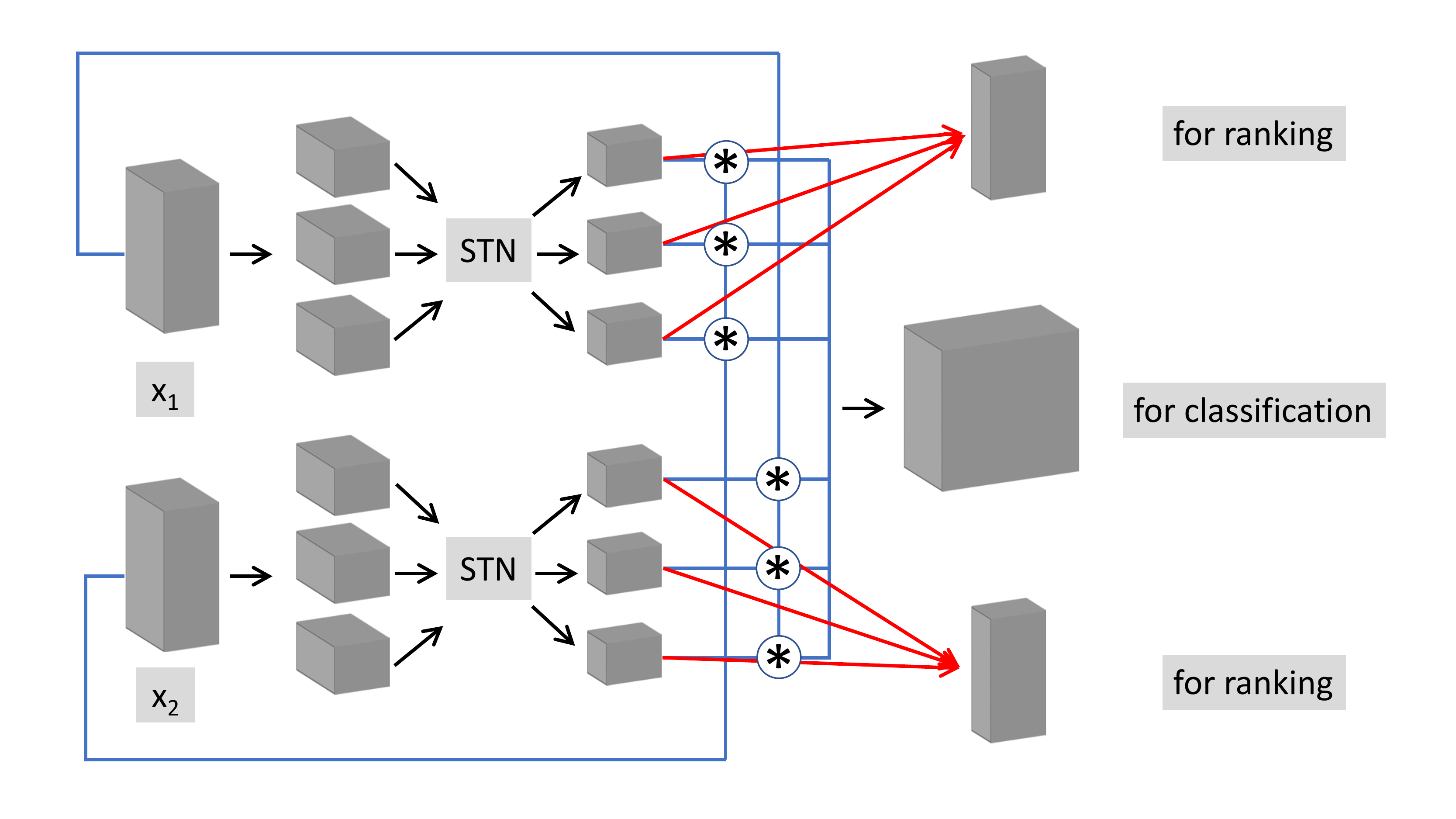}}
	\caption{The framework of CSN. * denotes depth-wise convolution. All the results of the depth-wise convolution are concatenated together as the outputs. The other outputs indexed by the red arrow are the feature maps extracted by the STN, which are used as the inputs for the ranking net.
	}
	\label{fig3}
\end{figure}

\noindent\textbf{Convolution similarity network} (CSN) is proposed to measure the similarity of two inputs. The framework of CSN is shown in Figure \ref{fig3}. Given feature maps of two images, we propose
efficient comparison with CSN: first, meaningful local regions are extracted by STNs; second, the local parts are treated as filters and thus the correlation between two groups of feature maps are computed in a more efficient way with fully convolutional structure. We will describe these two stages in details.

To find the meaningful contents from a pedestrian image is of great importance and challenging due to the large view point variation and occlusion. Spatial Transform Networks (STNs) are proven to be effective for images containing one kind of objects, which suits our application well. Therefore, we decide to use STN in our network to integrate the spatial attention.
%Previous work focuses on the rigid part in feature maps. For example, \cite{varior2016gated} just explores the possibility that giving different weight to different horizontal stripes in feature maps. Obviously, the irrelevant background in this case will have unmatched weight. \cite{Zhao_2017_CVPR} deploys RPN to make region proposals containing certain body parts. Nevertheless, the RPN needs to be trained on another dataset labeled with body parts, which might be a problem when the images in Person ReID datasets have quite different distributions. Similar to our work, \cite{Li_2017_CVPR} also proposes to use STNs for spatial attention. They learn the transformation parameters with STNs under different constraints for different parts of the input feature maps. In contrast, we reduce the number of constraints by dividing the feature maps directly.   

There are three components in an STN. The localization net learns the transformation parameters. We consider affine transformation which has 6 parameters. The grid generator and sampler together samples the input image and generates a new image with bilinear interpolation. In this case, the transformation is 

\begin{equation}\label{eq1}
\left(\begin{array}{c}   
w^{in} \\   
h^{in}   
\end{array}\right) =   
\left(\begin{array}{cccc}   
s_w &   r_w   & t_w \\   
r_h &   s_h   & t_h \\     
\end{array}\right)
\left(\begin{array}{c}   
w^{out} \\   
h^{out} \\
1   
\end{array}\right) 
\end{equation}
where $w_{in}$ and $h_{in}$ are the normalized pixel coordinates for input image and $w_{out}$ and $h_{out}$ are the normalized pixel coordinates for output image along the width and height. $s$, $r$ and $t$ are the scale, rotation, translation parameters. We suggest readers to have more details from \cite{jaderberg2015spatial}.

We have two fully convolutional STNs, $\mbox{STN}_2$ and $\mbox{STN}_3$ in our model for the $x_i^{(2)}$ and $x_i^{(3)}$ respectively. The structures of their localization nets are the same, shown in Table \ref{tab1}, but the weight parameters are not shared. A linear embedding with dimension 6 and hyperbolic tangent activation function are followed to output the 6 transformation parameters. We found that it was difficult to find the relatively important part in a global scale from $x_i^{(j)}$ by STNs. Therefore, $x_i^{(j)}$ are divided into three parts, namely the upper, the middle and the bottom, which are overlapped each other to some extent. The overlapping between two adjacent parts makes sure that the meaningful local visual features are covered. Note that all the three parts are sharing the same localization net. 

The size of the outputs of samplers is set to be $f_2\times f_2$ for $x_i^{(2)}$ and $f_3\times f_3$ for $x_i^{(3)}$. The value of $f_2$ is set to be larger than $f_3$ because the receptive fields for elements on $x_i^{(3)}$ are larger than those on $x_i^{(2)}$. Let $x_i^{(j, upper)}$, $x_i^{(j, middle)}$, $x_i^{(j, bottom)}$ denote the outputs of the three parts from $x_i^j$. Each of them is with the size $f_2\times f_2\times 96$ for $j=2$ and $f_3\times f_3\times 96$ for $j=3$. Now we have found the meaningful part of given feature maps. To search the corresponding similar features in another image, we treat the extracted parts as filters, slide them all over the feature maps of another image with stride $1$, which is like what a convolutional layer does. The similarity, modeled as cross-correlation between them, is described as
\begin{equation}\label{eq5}
s_i^{(j, part)} = x_i^{j} * x_{i'}^{(j, part)}
\end{equation}
Here $i, i' \in \{1, 2\}$ are different, indicating two input images. $part$ can be $upper, middle, bottom$. $*$ denotes depth wise convolution. This step can be implemented efficiently in existing deep learning frameworks, which gets rid of sampling the rigid parts from feature maps and comparing them with another mechanically. We choose depth wise convolution instead of traditional one due to the fact that different feature maps contain different activation patterns. Since the signal $x_i^{j}$ to be convolved and the filter $x_{i'}^{(j, part)}$ have the same number of channels and $x_i^{j}$ are padded with zero, $s_i^{(j, part)}$ have the same size as $x_i^{j}$. In order to be symmetric and fully exploit the similarity, depth wise convolution is performed between $x_2^{(j, part)}$ and $x_1^{(j)}$ as well as $x_1^{(j, part)}$ and $x_2^{(j)}$. Now for $j=2, 3$, we have 6 groups of $s_i^{(j, part)}$ respectively and we concatenate all the 6 groups along the channel direction and further do $2\times2$ maxpooling for $j=2$ to reduce noise and redundancy. The results are then the comprehensive similarity score maps between $x_1^{(j)}$ and $x_2^{(j)}$, denoted as $sim_j$. 

\noindent\textbf{Combination of visual similarities from different levels}. 
%When we compare two similar images, we will focus on different level visual similarities. If two persons in different images both wear a hat, which is a high level visual similarity, we will continue to compare the shape, color of the hats, which would determine the low level similarity. Based on the combination of these information, we will finally decide whether two images are matched or not. Our model also manage to simulate this process. 
It is well known that bottom convolutional layers in deep ConvNets contain low level features such as color, shape, texture, \etc, and higher convolutional layers learn  complex and semantic information.  In our case, we take the second and the third convolutional layers into account. $sim_2$ focuses on the low level visual similarity while $sim_3$ focuses on the relatively higher level visual similarity. $sim_2$ and $sim_3$ are concatenated together along the channel direction. Since there are 12 groups of depth wise convolution in total, now the size of the similarity score maps is $10\times4\times1152$, where $1152=96\times12$. These similarity score maps contain comprehensive information for the final decision.

3 convolutional layers(conv4, conv5, conv6 in Figure \ref{fig2}) are followed to process the similarity score maps. Specifically, $1\times1$ convolution is used to reduce the number of channels first. GAP is used to replace the fully connected layer to keep the fully convolutional structure. 

\noindent\textbf{Objective function} used to train the network is the combination of classification and ranking. Softmax loss is the objective function for binary classification.

\begin{equation}\label{eq6}
L_{cls} = \frac{1}{m}\sum_{i=1}^{m}[(1-y)p(y=0\vert \{x_1, x_2\})] + yp(y=1\vert \{x_1, x_2\})
\end{equation}
where $y=1$ when the input images are matched and $y=0$ otherwise. $p(y\vert \{x_1, x_2\})$ is the probability distribution of $y$ given input $\{x_1, x_2\}$, computed by softmax function. $m$ is the mini batch size.

The binary classification objective function intends to train a high accuracy model which somehow ignores the correct ranking. The combination of binary classification and contrastive loss may alleviate this issue and improve the performance substantially, which has been observed by \cite{Wang_2016_CVPR,chen2017multi}. However, their computation of contrastive or triplet loss depends on the global descriptor of the whole image. We argue that the global descriptors are not ideal for ranking task since they do not highlight the more discriminative parts of the original images. This issue can be resolved by our model. For the matched image pairs, it is reasonable to believe that they have the similar extracted parts, namely, local visual features. Given $x_i^{(j, upper)}$, $x_i^{(j, middle)}$, $x_i^{(j, bottom)}$ with spatial attention, we propose a ranking net to compute the ranking loss, which only consists of 3 convolutional layers. $x_i^{(j, upper)}$, $x_i^{(j, middle)}$, $x_i^{(j, bottom)}$ are firstly go through a convolutional layer with 96 filters of size $3\times3$ and a max-pooling layer. Then the $x_i^{(j, upper)}$, $x_i^{(j, middle)}$, $x_i^{(j, bottom)}$ are concatenated together for each $i$ along the vertical direction as they are extracted from different horizontal stripes. After another convolutional layer with 96 filters of size $3\times3$, the feature maps from the different layers, indexed by $j$, are concatenated again to form the descriptors for attended local parts. Then we also use GAP and linear embedding to obtain a 256 dimensional vector to represent the attended parts of one input image. The vectors for two images are $L_2$ normalized to make them comparable. Contrastive loss is computed for the two input images, 

\begin{equation}\label{eq7}
L_{ctr} = \frac{1}{2m} * \sum_{i=1}^{m}[yd^2 + (1-y)\mbox{max}(0, \alpha - d)^2 ]
\end{equation}

\begin{equation}\label{eq8}
d = \Vert \mathbf{r_1} - \mathbf{r_2} \Vert
\end{equation}
where $\mathbf{r_1}$ and $\mathbf{r_2}$ are the representations for two input images. $d$ is the Euclidean distance. $\alpha$ is the margin set to be $1.0$ in this paper.  With the help of this contrastive loss, images with similar attended parts are pushed closer in the embedding space. Otherwise, they are pulled further. 

The whole network is trained end-to-end with the combination of mentioned losses.

\begin{equation}\label{eq9}
%L_{com} = L_{cls} + L_{ctr} + 0.01*(L_{scale} + L_{rot} + L_{in}) 
L_{com} = L_{cls} + L_{ctr} 
\end{equation}

During testing, one query image and one image from the gallery are fed into the network. The final similarity score is computed as 

\begin{equation}\label{eq10}
\mbox{SimiScore} = s_{softmax} + \lambda*\frac{1}{d+\epsilon}
\end{equation}
where $s_{softmax}$ is the matched probability computed by softmax function and $d$ is the Euclidean distance in Eq. \ref{eq8}. $\lambda$ is set to be $0.2$ empirically and $\epsilon$ is set to be a small value like $0.0001$. All the images in the gallery are ranked based on their final similarity scores.

\subsection{Discussion}
\noindent\textbf{Efficiency}. We keep the implementation efficiency in mind when designing the model. In \cite{subramaniam2016deep}, the rigid local parts are sampled mechanically and compared with a restricted region of another image. However, sampling a rigid part directly from a tensor is usually avoided in existing deep learning frameworks for efficiency since it needs to index the elements from tensors.
Meanwhile, in the ReID application, the local rigid parts may not cover the important visual features and are not invariant to scale, translation and rotation. In contrast, sampling local meaningful part is done by a fully convolutional STN in our network, which is more flexible and effective. The sampled parts play the role of filters in a traditional convolutional layer, which is compatible with current deep learning frameworks, thus the implementation being much easier.

%\begin{figure}[htb]
%	\centering
%	\centerline{\includegraphics[width=5.5cm]{fig_5}}
%	\caption{Visualization of the input images and similarity score maps. Query images are in the first column. Testing images in second column are unmatched and matched ones. The similarity score maps in column 3 and 4 are from CSN-2 and CSN-3 respectively.}
%	\label{fig4}
%\end{figure}

%\begin{figure}[htb]
%	\centering
%	\centerline{\includegraphics[width=5.5cm]{fig_viz1}}
%	\caption{Visualization of the input images and similarity score maps. Query images are in the first column. Testing images in second column are unmatched and matched ones. The similarity score maps in column 3 and 4 are from CSN-2 and CSN-3 respectively.}
%	\label{fig7}
%\end{figure}

\begin{figure}[htb]
	\centering
	\begin{minipage}[]{.8\linewidth}
		%\centering
		\centerline{\includegraphics[width=5.0cm]{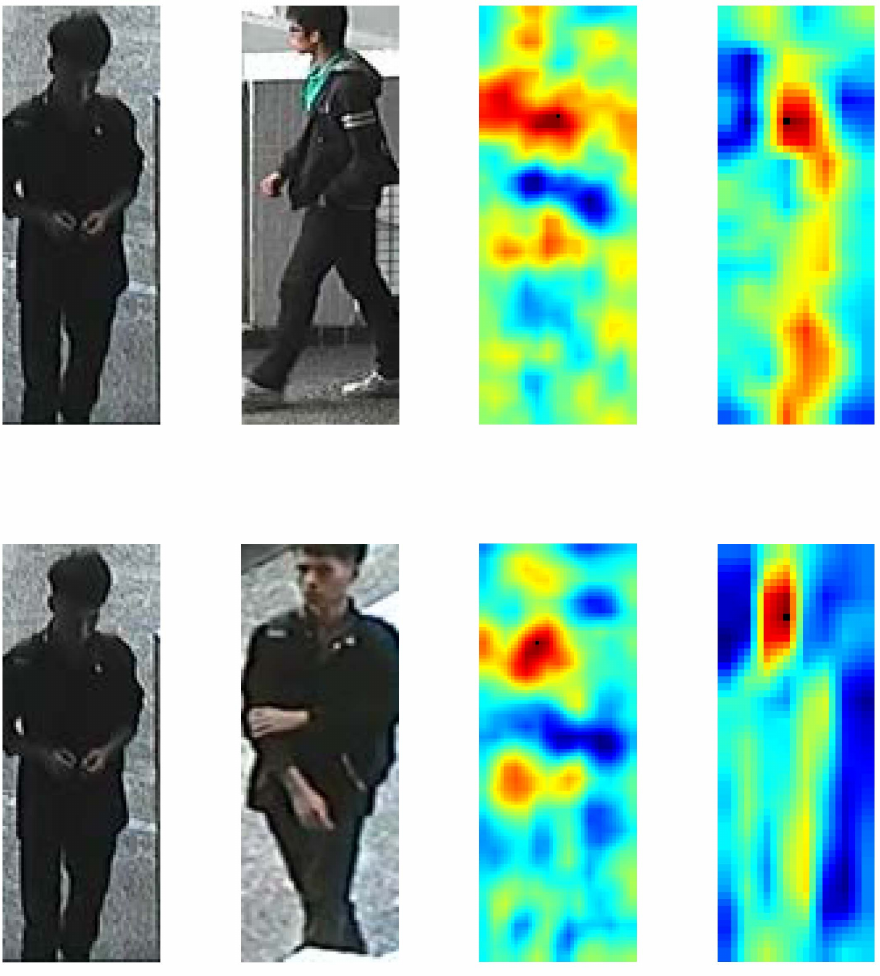}}
		\vspace{0.1cm}
		\centerline{(a)}\medskip
	\end{minipage}
	%\hfill
	\hspace{0.01cm}
	\begin{minipage}[]{.3\linewidth}
		%\centering
		\centerline{\includegraphics[width=5.0cm]{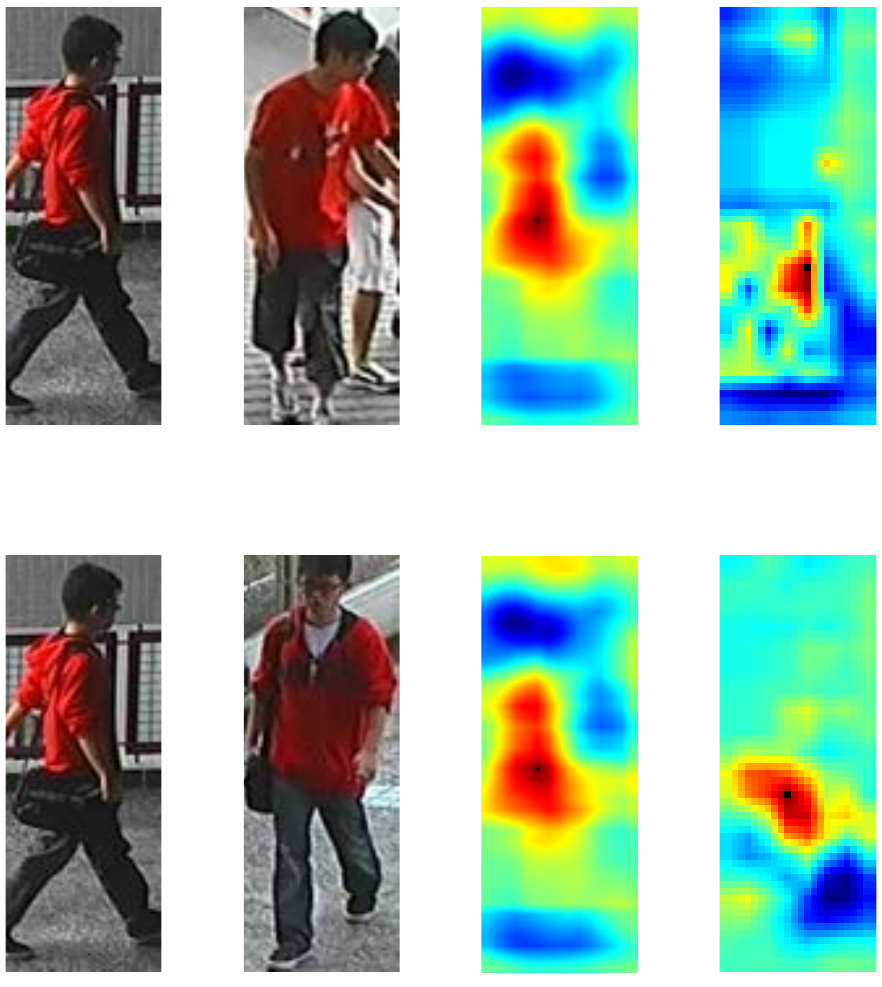}}
		%  \fbox{\includegraphics[width=4.0cm]{2}}
		\vspace{0.1cm}
		\centerline{(b)}\medskip
	\end{minipage}
	%\hfill
	\hspace{0.01cm}
	\caption{Visualization of the input images and similarity score maps. Query images are in the first column. Testing images in second column are unmatched and matched ones. The similarity score maps in column 3 and 4 are from CSN-2 and CSN-3 respectively.}
	\label{fig4}
\end{figure}

\noindent\textbf{Learned visual similarity from different levels}. To verify that our model indeed learns the visual similarities at different levels, we conduct some visualization experiments on the similarity score maps for dataset CUHK03 detected, shown in Figure \ref{fig4}. The similarity score maps in column 3 and 4 are the convolution results between the feature maps for query image and the attended upper and middle parts for the test image. For the positive pair in (a), it could be inferred that the similarity score map from CSN-2 focuses on the skin texture related features so that the exposed human skin parts like face and hands in the image get higher similarity scores, which again proves that STNs successfully grasp the meaningful local features. CSN-3, on the contrary, restricts its similarity on the face part only, which proves our assumption that CSN-2 and CSN-3 are dealing with low-level and complex semantic visual similarity. In (b), the negative testing image is quite confusing for the CSN-2 since the similarity score maps in the third column almost look the same because of the red tops. However, the similarity score maps from CSN-3 are easily distinguished due to the different similarity value in the bag area. It can be inferred that the similarity score maps from CSN-2, in this case, may mislead the model. However, the significant difference of the similarity score maps from CSN-3 will make sure that the model will give the right prediction. We can conjecture that when the high level visual similarities are confusing, the low level ones will help in turn. From the visualization, we can conclude that the combination of different level similarity is necessary for final success.

\noindent\textbf{Model extension}. There are two aspects to explore for model extension. On the one hand, we can have more CSNs in our network structure since our proposed CSN is fully differentiable and could be inserted in the network anywhere. We consider CSN-4 in experiments, which follows another convolutional layer after conv3 in Figure.\ref{fig2} and has the same structure as other CSNs. The CMC results on CUHK03 dataset in Table \ref{tab2} show that including more higher level visual similarity will indeed increase the performance by a large margin. On the other hand, we can achieve comparable performance with state-of-the-art methods leveraging pre-trained network, such as VGG \cite{simonyan2014very}, ResNet \cite{He_2016_CVPR}, \etc, in spite of simple three or four convolutional layers for feature extraction in our model. 
%It would be interesting to see how it affect the performance if we replace the first several layers with more advanced pre-trained network. We leave this as future work.
%man: this should have been done

%\begin{table}[!htbp]
%	\footnotesize
%	\centering
%	\caption{CMC results on CUHK03 dataset for our original model and extended model with CSN-4.}
%	\label{tab2}
%	\begin{tabular}{|c|c|c|c|c|c|c|}
%		\hline
%		\multirow{2}{*}{Method} & \multicolumn{3}{c|}{CUHK03 detected}             & \multicolumn{3}{c|}{CUHK03 labeled} \\ \cline{2-7} 
%		& top-1          & top-5          & top-10         & top-1      & top-5     & top-10     \\ \hline
%		ours-original           & 79.45          & 94.70          & 97.90          & 80.30      & 97.10     & 98.35      \\ \hline
%		ours-expanded             & \textbf{86.45} & \textbf{97.50} & \textbf{99.10} &            &           &            \\ \hline
%	\end{tabular}
%\end{table}

\section{Experiments}

% Please add the following required packages to your document preamble:
% \usepackage{multirow}
\begin{table*}[!htbp]
	\footnotesize
	\centering
	\caption{The CMC results comparison between our method and other state-of-the-art methods. }
	\label{tab2}
	\begin{tabular}{|c|c|c|c|c|c|c|c|c|c|c|c|c|}
		\hline
		\multirow{2}{*}{Method} & \multicolumn{3}{c|}{CUHK03 detected}             & \multicolumn{3}{c|}{CUHK03 labeled}              & \multicolumn{3}{c|}{CUHK01} & \multicolumn{3}{c|}{VIPeR}                       \\ \cline{2-13} 
		& top-1          & top-5          & top-10         & top-1          & top-5          & top-10         & top-1   & top-5   & top-10  & top-1          & top-5          & top-10         \\ \hline
		FPNN                    & 19.89          & 48.70          & 64.79          & 20.65          & 51.50          & 68.50          & 27.87   & 64.50   & 73.46   & -              & -              & -              \\
		ImpCNN                  & 44.96          & 76.50          & 83.47          & 54.74          & 87.80          & 93.88          & 65.00   & 89.00   & 93.12   & -              & -              & -              \\
		Joint                   & 52.17          & 85.30          & 91.20          & -              & -              & -              & 71.80   & 90.00   & 93.50   & 35.76          & 66.70          & 84.50          \\
		SiameseLSTM             & 57.30          & 80.10          & 88.30          & -              & -              & -              & -       & -       & -       & 42.40          & 68.70          & 79.40          \\
		S-CNN                   & 68.10          & 88.10          & 94.60          & -              & -              & -              & -       & -       & -       & 37.80          & 66.90          & 77.40          \\
		BDLatPart               & 67.99          & 91.04          & 95.36          & 74.21          & 94.33          & 97.54          & -       & -       & -       & -              & -              & -              \\
		ImpTriplet              & -              & -              & -              & -              & -              & -              & -       & -       & -       & 47.80          & \textbf{74.70} & 84.80          \\
		X-Corr                  & 72.04          & 92.10          & 96.00          & 72.43          & 92.50          & 95.51          & 81.23   & 95.00   & 97.39   & -              & -              & -              \\
		Quadruplet              & 75.53          & 95.15          & \textbf{99.16} & -              & -              & -              & 81.00   & 96.50   & 98.00   & 49.05          & 73.10          & 81.96          \\
		DGD                     & -              & -              & -              & 72.58          & 91.59          & 95.21          & 66.60   & -       & -       & 38.6           & -              & -              \\
		MTDNet                  & 74.68          & 95.99          & 97.47          & -              & -              & -              & 78.50   & 96.50   & 97.50   & 47.47          & 73.10          & 82.59          \\
		MuDeep                  & 75.64          & 94.36          & 97.46          & 76.87          & 96.12          & 98.41          & 79.01   & 97.00   & 98.96   & 43.03          & 74.36          & \textbf{85.76} \\ \hline
		DPFL                    & 82.00          & -              & -              & 86.70          & -              & -              & -       & -       & -       & -              & -              & -              \\
		DeepAlign               & 81.60          & 97.30          & 98.40          & 85.40          & 97.60          & 99.40          & \textbf{88.50}   & \textbf{98.40}   & \textbf{99.60}   & 48.70          & \textbf{74.70} & 85.10          \\
		PDC                     & 78.29          & 94.83          & 97.15          & \textbf{88.70} & \textbf{98.61} & 99.24          & -       & -       & -       & 51.27          & 74.05          & 84.18          \\
		Spindle                 & -              & -              & -              & 88.50          & 97.80          & 98.60          & 79.90   & 94.40   & 97.10   & \textbf{53.80} & 74.10          & 83.20          \\
		JLML                    & 80.60          & 96.90          & 98.70          & 83.20          & 98.00          & 99.40          & -       & -       & -       & 50.20          & 74.20          & 84.30          \\ \hline
		Ours-(L2, L3)           & 79.45          & 94.70          & 97.90          & 80.30          & 97.10          & 98.35          & 86.55   & 97.70   & 98.70   & 48.03          & 72.90          & 82.15          \\
		Ours-(L2, L3, L4)           & \textbf{86.45} & \textbf{97.50} & 99.10          & 87.50          & 97.85          & \textbf{99.45} &  88.20       &   98.20      &   99.35      &   50.10         &    73.10        &     84.35           \\ \hline
	\end{tabular}
\end{table*}

\subsection{Datasets and evaluation metrics}
We test our model on four dataset: CUHK03 detected and labeled \cite{li2014deepreid}, CUHK01 \cite{li2012human}, VIPeR \cite{gray2007evaluating}. CUHK03 is a large dataset containing 13,164 images fro 1,360 identities captured by 6 cameras. This dataset has two kinds of pedestrian boxes: detected by algorithms and labeled manually, both of which we will use.  Following the setting as \cite{li2014deepreid}, we randomly choose 1160 identities for training, 100 for validation and 100 for testing. CUHK01 is a middle size dataset containing 3884 images of 971 identities. For our experiments, we follow the setting as \cite{subramaniam2016deep} and randomly choose 871 identities for training and 100 for testing. VIPeR is a small size datasets with 632 identities, for which we randomly choose half of them for training and half for testing.

Cumulative Matching Characteristics (CMC) are reported to evaluate the performance. There are only one query image and one matched image in the gallery for each testing identity, \ie, single shot setting. Rank-$k$ accuracy stands for the accuracy that the matched image in the gallery is included in the top-$k$ answers based on the similarity score.

\subsection{Implementation details}
We implement our model with TensorFlow \cite{abadi2016tensorflow}. ADAM \cite{kingma2014adam} is used to optimize the network with learning rate $0.0005$. We train the network for 5 epochs. Weighting decay is set to be $0.0005$ to avoid over-fitting. Batch normalization \cite{ioffe2015batch} is used to make the training stable and fast to converge. The mini batch size is set to be 256 for CUHK03 and 128 for other two datasets. $f_1$ and $f_2$ are set to be $10$ and $5$ respectively. Data augmentation is also adopted for training as \cite{Ahmed_2015_CVPR} and \cite{subramaniam2016deep}. We randomly sample 2 images for CUHK03 and 5 for others from the original image center and also flip it horizontally. On the other hand, since the negative pairs in the training set outnumber the positive pairs significantly, the model easily falls into over-fitting and predicts all the pairs as negative. Therefore we only randomly choose two negative pairs for each positive pair. Note that we do not introduce hard negative mining, which simplifies the training process. We use one NVIDIA TitanX GPU to train the model. During inference, the model takes 1250 pairs of images as input and obtains the final score in about 1.6s.

Empirically, we found that learning the transformation parameters without constraints would cause several issues, such as negative scale parameters, falling out of the original image and so on, which are also pointed out by \cite{Li_2017_CVPR}. Therefore, some similar prior constraints are put on the 6 transformation parameters, which are used to keep the scale parameters larger than 0, which avoids the upside down case, and to keep the results staying in the original images. In addition, we simply use $L_1$ regularization for $r_w$ and $r_h$ since rotation is seldom happened in real world cases. As we discussed before, we divide the feature maps into three parts for the Localization Net. In particular, the upper part of $x_i^{(2)}$ is composed of the row 1 to 20 of $x_i^{(2)}$, middle row 10 to 30, bottom row 20 to 40. Similarly, the upper part of $x_i^{(3)}$ is composed of the row 1 to 10 of $x_i^{(3)}$, middle row 5 to 15, bottom row 10 to 20.  

%\cite{Li_2017_CVPR} constrain the translation parameters $t_w$ and $t_h$ so that the extracted parts are located near the center of the input feature maps. We do not have to deploy this constraint here because we divide the input feature maps into three overlapped parts and the meaningful contents we want to extract may not exist in the center of each part. 

\subsection{Comparison with state-of-the-arts}

We compare our approach with several state-of-the-art methods in recent years, including FPNN \cite{li2014deepreid}, ImpCNN \cite{Ahmed_2015_CVPR}, SiameseLSTM \cite{varior2016siamese}, S-CNN \cite{varior2016gated}, BDLatPart \cite{Li_2017_CVPR}, MTDNet \cite{chen2017multi}, X-Corr \cite{subramaniam2016deep}, Quadruplet \cite{Chen_2017_CVPR}, ImpTriplet \cite{Cheng_2016_CVPR}, DGD \cite{xiao2016learning}, Joint \cite{Wang_2016_CVPR}, MuDeep \cite{Qian_2017_ICCV}, Spindle \cite{Zhao_2017_CVPR}, DPFL \cite{Chen_2017_ICCV_Workshops}, DeepAlign \cite{Zhao_2017_ICCV}, PDC \cite{Su_2017_ICCV} and JLML \cite{ijcai2017-305}. In particular, we want to compare our method with MTDNet, X-Corr and BDLatPart. MTDNet combines binary classification and ranking together with the global descriptors alone for input images. X-Corr is a Siamese network computing the similarity as correlation between rigid parts of input images. BDLatPart introduces the STN into their structure to extract local meaningful parts. However, BDLatPart only considers to learn the representation for one input image, which significantly differs from our work. The results are shown in Table \ref{tab2}. The methods in the upper part train their models from scratch on the ReID datasets alone while the methods in the middle part either train a sub-network on another datasets for more supervision or use pre-trained networks like Inception-V3 as backbone structure. As we could guess, the middle methods usually outperforms the upper ones with the help of pre-training or more information. We implemented two models, one is the same as Figure \ref{fig2} denoted as Ours-(L2, L3) since it considers visual similarities from the second and the third convolutional layers and the other including one more CSN to consider higher level visual similarity as discussed in Section 3.2, denoted as Ours-(L2, L3, L4).

%Spindle \cite{Zhao_2017_CVPR} achieves the state-of-the-art results on CUHK03 with top-1 accuracy $=88.5$. Note that we do not present it here for comparison since it uses another dataset to train the RPN, which might make the comparison unfair. 

%Spindle employing RPN achieves best results for CUHK03 detected dataset and VIPeR dataset, which confirms the effectiveness of the introduction of spatial attention. However, Spindle has to train the RPN on another dataset. 

%Our approach outperforms other methods on CUHK03 detected and labeled datasets as well as CUHK01 dataset and achieves competing result on VIPeR. For CUHK03 detected, if we consider the model trained only on the CUHK03 detected dataset from scratch, our results are the state-of-the-art. For CUHK03 labeled and CUHK01 datasets, our method again outperforms others by $5\% \sim 6\%$ margin. In particular, as we expected, X-Corr employing correlation between rigid parts of input images suffers from the restricted comparing region and BDLatPart leveraging STN suffers from the absence of comparison. Both of them obtain inferior results than our model for all datasets, which convinces us to believe that our model is the better way to compare the input images with the usage of STNs.

Our extended model achieves the best top-1 accuracy for CUHK03 detected datasets, outperforming all the methods by a large margin. On the other three datasets, our extended model can get performance comparable to state of the art methods with smaller model size and computation amount. Even our weaker model, ours-(L2, L3), can beat all the methods in the upper table for CUHK03 and CUHK01 datasets. As expected, X-Corr suffers from the mechanical correlation computation and restricted comparing regions while BDLatPart fails to extract discriminative enough representations for each identity even with the help of STNs, which convinces us that our model has the better way to compute similarity effectively with the usage of STNs. The performance of MTDNet is also inferior to our approach due to the lack of explicit similarity computation and spatial attention on the discriminative local parts. The results shown here demonstrate the effectiveness of our similarity computation at multiple levels.

\subsection{Ablation analysis}

To further understand our model, we conduct several ablation experiments for our model on CUHK03 detected dataset, which contains the images more similar to the real world application.

First, we remove the contrastive loss $L_{ctr}$ function and train the network with the same settings carefully. The CMC results, denoted as ours-cls, are shown in Table \ref{tab3}. We can see that without the ranking loss, the performance degrades $3\%$ for the rank 1 accuracy. 
%Indeed, the ranking loss will provide additional information to the model as we expected. 
%For the model trained with combined loss in Eq. \ref{eq9}, the tradeoff factor $\lambda$ in Eq.\ref{eq10} between $s_{softmax}$ and $d$ is studied, shown in Figure \ref{fig4} with different value from $0$ to $1$. It can be observed that even when $\lambda = 0$, the top-1 accuracy, $77.2\%$, is higher than ours-cls, which proves that the ranking loss helps the binary classification converge to a better optimal than itself alone. The top-1 accuracy reaches the highest value when $\lambda=0.2$. When $\lambda$ keeps increasing, the top 1 accuracy decreases apparently, which implies that the information provided by ranking loss is likely to be less important than the binary classification. Therefore, we can conclude that the combination of two losses is necessary.

%\begin{figure}[!htb]
%	\centering
%	\centerline{\includegraphics[width=5.5cm]{fig_3}}
%	\caption{Top 1 accuracy on CUHK03 detected dataset with different $\lambda$.}
%	\label{fig5}
%\end{figure}

Then we investigate the importance of visual similarities at different levels of the proposed network. We keep only one CSN in the model and train the network under the same strategy. New models are denoted as Ours-L2, Ours-L3 and Ours-L4. The results are shown in table \ref{tab3}. As we discussed before, CSN-2 computes the low level visual similarity such as edge, shape, \etc while CSN-3 and CSN-4 focuses on the higher level similarity containing semantical information. We can find that the high level visual similarity is more important than the lower level one. The CSN-4 alone helps the model achieves similar performance to the combination of CSN-2 and CSN-3. However, all the models with single CSN obtain inferior performances to the one utilizing combination of 3 CSNs, indicating that low level similarity provides additional information ignored by the high level one and thereby the necessity of combing low and high level visual similarities.

\begin{table}[!htb]
	\centering
	\caption{The CMC results comparison between our original method and modified ones on CUHK03 detected dataset.}
	\label{tab3}
	\begin{tabular}{|c|c|c|c|}
		\hline
		Method                    & top-1          & top-5          & top-10                  \\ \hline
		Ours-cls                  & 75.90          & 94.55          & 97.85                   \\ \hline
		Ours-L2                    & 74.70          & 93.52          & 96.45              \\ \hline
		Ours-L3                    & 76.55          & 93.70          & 96.85              \\ \hline
		Ours-L4                    & 79.15          & 94.45          & 97.80              \\ \hline
		Ours-(L2, L3)             & 79.45 & 94.70 & 97.90                                      \\ \hline
		Ours-(L2, L3, L4)             & \textbf{86.45} & \textbf{97.50} & \textbf{99.10}          \\ \hline
	\end{tabular}
\end{table}

Last but not least, different configurations of the proposed network are studied. We examine the usefulness of dividing the images into three horizontal stripes in C1. In C2, we replace adaptive STN with fixed central cropping, i.e., we crop a center region with the same size as STN from each horizontal stripe. C3 is our proposed model with only Level 4 similarity and C4 is the original model with Multi Level similarities. Results on CUHK03 detected dataset are shown in Table \ref{tab4}. Comparing C1 and C4, we can observe that without dividing, it becomes difficult for STN to find the meaningful regions. In fact, the result of STN without dividing is worse than central cropping(C2). C2 achieves reasonable results when multi level similarities are used, which demonstrates the effectiveness of multi level similarities. With the combination of dividing and STN, we can compute more accurate similarity score maps from different feature levels, and this leads to the best performance as shown in C4.

\begin{table}[!htb]
	\footnotesize
	\centering
	\caption{CMC results for different configurations on CUHK03 detected datasets. ML here means Multi Level(L2, L3, L4).}
	\label{tab4}
	\begin{tabular}{|c|c|c|c|c|c|c|}
		\hline
		config. & dividing & STN & ML & top-1           & top-5           & top-10          \\ \hline
		C1       & $\times$        & \checkmark   & \checkmark  & 79.40          & 94.95          & 98.40          \\ \hline
		C2       & \checkmark    & $\times$   & \checkmark  & 82.00          & 96.40          & 98.55          \\ \hline
		C3       & \checkmark    & \checkmark   & $\times$  & 79.15          & 94.45          & 97.80          \\ \hline
		C4       & \checkmark    & \checkmark   & \checkmark  & \textbf{86.45} & \textbf{97.50} & \textbf{99.10} \\ \hline
	\end{tabular}
\end{table}

%Last but not least, the effect of regularization on the 6 transform parameters is studied. We consider three cases: 1) removing all the constraints; 2) ignoring the rotation situations by keep the $r_w$ and $r_h$ 0; 3) adopting regularization the same as \cite{Li_2017_CVPR}. The results on CUHK03 detected datasets are show in Table \ref{tab4}. Our method outperforms all the three cases. In (1), we found that the extracted parts fell out of the original images easily, which is definitely what we want to avoid. Serious rotation situation is rarely seen in real world application but rotating to a small extent is likely to exist due to the cameras, which is proven by the $2\%$ margin between our method and (2). If we adopt the same regularization as BDLatPart, the performance drops drastically in (3), which justifies the superiority of our dividing-and-regularizing strategy.     

%\begin{table}[!htb]
%	\centering
%	\caption{CMC results on CUHK03 detected datasets for different regularization on transform parameters.}
%	\label{tab4}
%	\footnotesize
%	\begin{tabular}{|c|c|c|c|}
%		\hline
%		Method & top-1          & top-5          & top-10       \\ \hline
%		1      & 77.25          & 94.35          & 96.65          \\ \hline
%		2      & 77.80          & 94.55          & 97.20          \\ \hline
%		3      & 72.10          & 93.00          & 97.10          \\ \hline
%		Ours   & \textbf{79.45} & \textbf{94.70} & \textbf{97.90} \\ \hline
%	\end{tabular}
%\end{table}

\subsection{Complexity Analysis}
We compare the proposed model with five recent proposed models in model size and computation complexity, which are measured by the number of parameters and the value of FLOPs during inference. X-Corr\cite {subramaniam2016deep} and BDLatPart \cite{Li_2017_CVPR} are trained from scratch so we estimate the number of parameters and FLOPs by ourselves. DPFL \cite {Chen_2017_ICCV_Workshops} and DeeAlign \cite{Zhao_2017_ICCV} use pre-trained Inception-V3 \cite{Szegedy_2016_CVPR} and GoogLeNet \cite{Szegedy_2015_CVPR} as their backbone structures, which are considered as main contributors for complexity. JLML \cite{ijcai2017-305}, based on ResNet39, discloses the complexity in their paper. Table \ref{tab5} shows that our original model has the smallest model size and least computation amount while outperforming X-Corr\cite{subramaniam2016deep} and BDLatPart\cite{Li_2017_CVPR} by a large margin ($8\%$ to $10\%$) on CUHK03 detected dataset. The performance of our extended model, ours-(L2, L3, L4), with the second lowest complexity, can be comparable to the methods with pre-trained networks. 

\begin{table}[htb]
	\centering
	\caption{Comparison of model size and complexity. $\#$param: number of parameters. M: Million. G: Giga.}
	\label{tab5}
	\begin{tabular}{|c|c|c|c|}
		\hline
		Model     & \#param(M)   & FLOPs(G)      & Depth       \\ \hline
		X-Corr    & 2.2          & 1.58          & \textbf{10} \\ \hline
		BDLatPart & 1.4          & 1.80          & 25          \\ \hline
		DPFL      & 35           & 6.00         & 40          \\ \hline
		JLML      & 7.2           & 1.54         & 39          \\ \hline
		DeepAlign      & 6           & 1.57         & 22          \\ \hline
		Ours-(L2, L3)      & \textbf{0.5} & \textbf{0.96} & \textbf{12}          \\ \hline
		Ours-(L2, L3, L4)      & \textbf{0.8} & \textbf{1.31} & 18          \\ \hline
	\end{tabular}
\end{table}

\section{Conclusion}

In this work, we propose a novel fully convolutional Siamese network for Person ReID. Our system extracts features from local parts of one input image and then computes the visual similarity with another input image through depth-wise convolution. By exploiting two or more CSNs at different convolutional layers, we obtain visual similarities at different levels. This approach avoids sampling the rigid parts of input images and could be implemented efficiently. We further enhance the system by considering contrastive loss based on descriptors for the extracted local parts. Extensive experiments on four Person Re-ID datasets show that our approach could achieve comparable performance with  recent state-of-the-art, at a lower computational complexity and model size.  Ablation and visualization experiments show that the visual similarities from different levels all contribute to the overall improvement. We also provide the comparison in model size and complexity and show that our method can achieve good performance at lower complexity. 

\section*{Acknowledgment}

This work was supported by both ST Electronics and the National Research Foundation(NRF), Prime Minister's Office, Singapore under Corporate Laboratory at University Scheme (Programme Title: STEE Infosec - SUTD Corporate Laboratory).

{\small
\bibliographystyle{ieee}
\bibliography{bibfile}
}

\end{document}